\documentclass{article}
\usepackage{template}

\usepackage[utf8]{inputenc} 
\usepackage[T1]{fontenc}    
\usepackage{url}            
\usepackage{booktabs}       
\usepackage{amsfonts}       
\usepackage{nicefrac}       
\usepackage{microtype}      
\usepackage{lipsum}
\usepackage{fancyhdr}       
\usepackage{graphicx}       
\graphicspath{{Fig/}}     
\usepackage{amsmath}
\usepackage{amssymb}
\usepackage{graphicx}
\usepackage{enumerate}
\usepackage{setspace}
\usepackage{comment}
\usepackage[utf8]{inputenc}
\usepackage[breaklinks=true]{hyperref}
\usepackage[noend]{algorithmic}
\usepackage{subfigure}
\usepackage{algorithm}
\usepackage[spanish, english]{babel}
\usepackage{url}
\usepackage{notoccite}
\usepackage{fancyhdr}
\usepackage{ragged2e} 
\usepackage[labelfont=bf, center]{caption}
\usepackage{multirow}
\usepackage{stackengine}
\usepackage{booktabs}
\usepackage[table,xcdraw]{xcolor}
\usepackage[normalem]{ulem}
\useunder{\uline}{\ul}{}
\usepackage{calrsfs}
\usepackage{textcomp}
\usepackage{bbm}
\usepackage{tikz}
\DeclareMathAlphabet{\pazocal}{OMS}{zplm}{m}{n}
\def\BibTeX{{\rm B\kern-.05em{\sc i\kern-.025em b}\kern-.08em
    T\kern-.1667em\lower.7ex\hbox{E}\kern-.125emX}}
\usetikzlibrary{shapes}
\usetikzlibrary{automata,positioning}

\title{SAVAE: Leveraging the variational Bayes autoencoder for survival analysis}

\author{
  Patricia A. Apellániz \\
  Information Processing \\and Telecommunications Center \\
  Universidad Politécnica de Madrid \\
  Madrid\\
  \textit{patricia.alonsod@upm.es} \\
  \And
  Juan Parras \\
  Information Processing \\and Telecommunications Center \\
  Universidad Politécnica de Madrid \\
  Madrid\\
  \textit{j.parras@upm.es} \\
  \And
  Santiago Zazo \\
  Information Processing \\and Telecommunications Center \\
  Universidad Politécnica de Madrid \\
  Madrid\\
  \textit{santiago.zazo@upm.es} \\
}

\begin{document}

\maketitle

\justifying

\begin{abstract}
As in many fields of medical research, survival analysis has witnessed a growing interest in the application of deep learning techniques to model complex, high-dimensional, heterogeneous, incomplete, and censored medical data. Current methods often make assumptions about the relations between data that may not be valid in practice. In response, we introduce SAVAE (Survival Analysis Variational Autoencoder), a novel approach based on Variational Autoencoders. SAVAE contributes significantly to the field by introducing a tailored ELBO formulation for survival analysis, supporting various parametric distributions for covariates and survival time (as long as the log-likelihood is differentiable). It offers a general method that consistently performs well on various metrics, demonstrating robustness and stability through different experiments. Our proposal effectively estimates time-to-event, accounting for censoring, covariate interactions, and time-varying risk associations. We validate our model in diverse datasets, including genomic, clinical, and demographic data, with varying levels of censoring. This approach demonstrates competitive performance compared to state-of-the-art techniques, as assessed by the Concordance Index and the Integrated Brier Score. SAVAE also offers an interpretable model that parametrically models covariates and time. Moreover, its generative architecture facilitates further applications such as clustering, data imputation, and the generation of synthetic patient data through latent space inference from survival data.
\end{abstract}


\section{Introduction}
In recent years, there has been a significant transformation in medical research methodologies towards the adoption of Deep Learning (DL) techniques for predicting critical events, such as disease development and patient mortality.  Despite their potential to handle complex data, practical applications in this domain remain limited, with most studies still relying on traditional statistical methods.

Survival Analysis (SA), or time-to-event analysis, is an essential tool for studying specific events in various disciplines, not only in medicine but also in fields such as recommendation systems \cite{sa_rec}, employee retention \cite{sa_job}, market modeling \cite{sa_market}, and financial risk assessment \cite{sa_fin}.

According to the existing literature, the Cox proportional hazards model (Cox-PH) \cite{cph} is the dominant SA method that offers a semiparametric regression solution to the non-parametric Kaplan-Meier estimator problem \cite{kaplanmeier}. Unlike the Kaplan-Meier method, which uses a single covariate, Cox-PH incorporates multiple covariates to predict event times and assess their impact on the hazard rate at specific time points. However, it is crucial to acknowledge that the Cox-PH model is built on certain strong assumptions. One of these is the proportional hazards assumption, which posits that different individuals have hazard functions that remain constant over time. Furthermore, the model assumes a linear relation between the natural logarithm of the relative hazard (the ratio of the hazard at time $t$ to the baseline hazard) and the covariates. Furthermore, it assumes the absence of interactions among these covariates. It is worth noting that these assumptions may not hold in real-world datasets, where complex interactions between covariates and non-linear relations might exist. Other traditional parametric statistical models for SA make specific assumptions about the distribution of event times. For instance, models like those presented in \cite{lee_wang, ranganath_2016} assume exponential and Weibull distributions, respectively, for event times. However, one drawback of these models is that they lack flexibility when it comes to changing the assumed distribution for survival times, making them less adaptable to diverse datasets.

In response, researchers have explored Deep Neural Networks (DNNs) to effectively capture the intricate and non-linear relations between predictive variables and a patient's risk of failure. Significant emphasis has been placed on improving the Cox PH model, which has been the standard approach in SA.

Recent approaches have introduced Neural Networks (NN) in various configurations, either enhancing the Cox-PH model with neural components or proposing entirely novel architectures. This exploration of NN applications for SA traces back to 1995 with the work of \cite{faraggi}, who initially employed a simple feed-forward NN to replace linear interaction terms while incorporating non-linearities. Subsequently, the field saw the emergence of DeepSurv \cite{deepsurv}, a model designed to extract non-linearities from input data, albeit still assuming the proportional hazards assumption. This assumption persists in other related models like the one proposed by \cite{luck}. Beyond addressing non-linearity, some researchers have sought to enhance prediction accuracy and model interpretability by combining Bayesian networks with the Cox-PH model, as demonstrated by \cite{kraisangka}. Additionally, efforts have been made to introduce concepts that facilitate analysis when data availability is limited, as seen in the work of \cite{vinzamuri_reddy, vinzamuri}. However, it is essential to note that all these models still depend on the proportional hazards assumption. As a result, novel architectures such as DeepHit \cite{deephit} have emerged as alternatives that do not rely on the proportional hazards assumption. While DeepHit has exhibited superior performance compared to other state-of-the-art models, it operates exclusively in the discrete-time domain, which comes with certain limitations, notably the requirement for a dataset with a substantial number of observations, a condition that may not be feasible in real-world scenarios.

In light of the persistent limitations of existing approaches in the realm of SA, this paper introduces a novel, versatile algorithm grounded in DL advances, named SAVAE (Survival Analysis Variational Autoencoder). SAVAE has been meticulously designed to predict the time distribution that leads to a predefined event and adapts to application in various domains, with a specific focus on the medical context. Then, our main contributions consist of:
\begin{itemize}
    \item We introduce a generative approach that underpins the development of a flexible tool, SAVAE, based on Variational Autoencoders (VAEs). SAVAE can effectively reproduce the data by analytically modeling the discrete or continuous time to a specific event. This analytical approach enables the calculation of all necessary statistics with precision, as the output provided by SAVAE are the parameters of the predicted time distribution
    \item SAVAE is a flexible tool that enables us to use a wide variety of distributions to model the time-to-event and the covariates. This allows us to not assume proportional hazards. By using NN, it permits modeling complex, non-linear relations between the covariates and the time-to-event too, as opposed to the linearity assumptions in the state of the art. Also, the time-to-event is trained with standard likelihood techniques, unlike state-of-the-art models like DeepHit, which trains the Concordance Index (C-index). This makes our approach more general and flexible, as any differentiable distribution could be used to model the time and the covariates.
    \item Furthermore, our proposal can be trained on censored data, effectively leveraging information from patients who have not yet experienced the event of interest.
    \item We have conducted comprehensive time-to-event estimation experiments using datasets characterized by continuous and discrete time-to-event values and varying covariate natures, encompassing both clinical and genomic data. These experiments involve a comparative analysis with the traditional Cox-PH model and other DL techniques. The results indicate that SAVAE is competitive with these models in terms of the C-index and the Integrated Brier score (IBS).
\end{itemize}

\section{Background}\label{background}
To establish context, we will define SA and VAEs. SA is a branch of applied statistics that examines random processes related to system failures and mortality. Following this, we will provide an analytical overview of VAEs before introducing SAVAE.

\subsection{Survival Analysis}\label{sa}
In a conventional time-to-event or SA setup, \textit{N} observations are given. Each of these observations is described by $D = (x_i, t_i, d_i)^N_{i=1}$ triplets, where $x_i=(x_i^1,..., x_i^L)$ is an $L$-dimensional vector where $l=1,2,...,L$ indexes the covariates, $t_i$ is the time-to-event, and $d_i \in \{0,1\}$ is the censor indicator. When $d_i = 0$ (censored), the subject has not experienced an event up to time $t_i$, while $d_i = 1$ indicates the observed events (ground truth). SA models are conditional on covariates: time probability density function $p(t\vert x)$, hazard rate function (the instantaneous rate of occurrence of the event at a specific time) $h(t\vert x)$, or survival function $S(t\vert x) = P(T>t) = 1-F(t\vert x)$, also known as the probability of a failure occurring after time $t$, where $F(t\vert x)$ is the Cumulative Distribution Function (CDF) of the time. From standard definitions of the survival function, the relations between these three characterizations are formulated as:
\begin{equation}\label{eq:sa}
    p(t\vert x) = h(t\vert x)S(t\vert x).
\end{equation}

\subsection{Vanilla Variational Autoencoder}
In 2013, \cite{kingma} proposed the original VAE, a powerful approach employing DNNs for Bayesian inference. It addresses the problem of a dataset consisting of $N$ i.i.d. samples $x_i$ of a continuous or discrete variable, where $i \in 1,2, ..., N$, $x_i$ are generated by the following random process, which is depicted in Figure \ref{fig:vae}:
\begin{figure}[!t]
    \centering
    \includegraphics[width=0.35\textwidth]{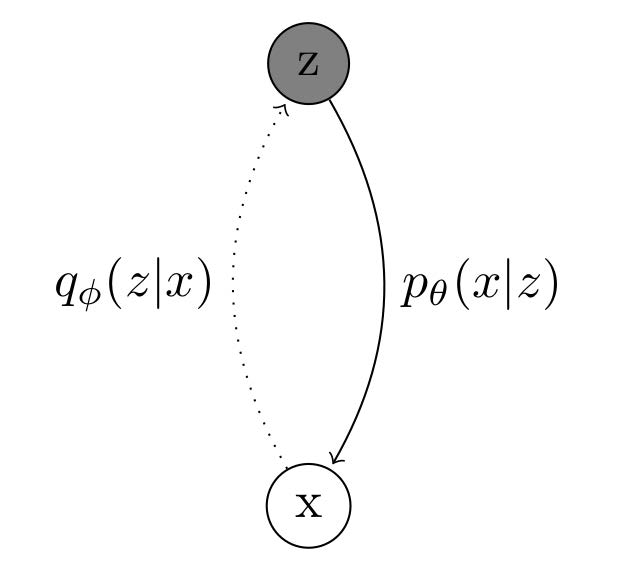}
    \caption{Bayesian VAE vanilla model. The shaded circle refers to the latent variable $z$, and the white circle refers to the observable $x$. Probabilities $p_\theta(x\vert z)$ and $q_\phi(z\vert x)$ denote, respectively, the generative model and the variational approximation to the posterior, since the true posterior $p(z\vert x)$ is unknown.}
    \label{fig:vae}
\end{figure}

\begin{enumerate}
    \item A latent variable $z_i$ is sampled from a given prior probability distribution $p(z)$. \cite{kingma} assumes a form $p_\theta(z)$, i.e., the prior depends on some parameters $\theta$, but its main result drops this dependence. Therefore, in this paper, a simple prior $p(z)$ is assumed.
    \item A conditional distribution, $p_\theta(x\vert z)$, with parameters $\theta$ generates the observed values, $x_i$. This process is governed by a generative model. Certain assumptions are made, including the differentiability of probability density functions (pdfs), $p(z)$, and $p_\theta(x\vert z)$, regarding $\theta$ and $z$.
\end{enumerate}

The latent variable $z$ and the parameters $\theta$ are unknown. Without simplifying assumptions, evaluating the true posterior density $p_\theta(x) = \int p(z)p_\theta(x\vert z)dz$ is infeasible.  This true posterior density can be defined as Equation \ref{eq:bayes} using Bayes' theorem:
\begin{equation}
    p_\theta(z\vert x) = \frac{p_\theta(x\vert z)p(z)}{p_\theta(x)}.
    \label{eq:bayes}
\end{equation}

Variational methods offer a solution by introducing a variational approximation, $q_\phi(z\vert x)$, to the true posterior. This approximation involves finding the best parameters for a chosen family of distributions through optimization. The quality of the approximation depends on the expressiveness of this parametric family.
\subsubsection{ELBO derivation}\label{elbo}
Since an optimization problem must be solved, the optimization target needs to be developed. Considering $x_i$ are assumed to be i.i.d., the marginal likelihood of a set of points $\{x_i\}_{i=1}^{N}$ can be expressed as 
\begin{equation}
    \log p_\theta(x_1, x_2, ..., x_N) = \sum_{i=1}^{N}\log p_\theta(x_i),
\end{equation}
where 
\begin{equation}
    \begin{split}
        p_\theta(x) = \int p_\theta(x,z)dz = \int p_\theta(x,z) \frac{q_\phi(z\vert x)}{q_\phi(z\vert x)}dz
        = \mathbb{E}_{q_\phi(z\vert x)} \left[\frac{p_\theta(x,z)}{q_\phi(z\vert x)}\right].
    \end{split}
\end{equation}
Using Jensen's inequality, we can obtain:
\begin{equation}
    \begin{split}
        \log p_\theta(x) = \log \Bigg[\mathbb{E}_{q_\phi(z\vert x)}\left[\frac{p_\theta(x,z)}{q_\phi(z\vert x)}\right]\Bigg] \ge \mathbb{E}_{q_\phi(z\vert x)}\left[ \log\frac{p_\theta(x,z)}{q_\phi(z\vert x)}\right].
    \end{split}
    \label{eq:jensen}
\end{equation}
Rearranging Equation \ref{eq:jensen}, we can express it as follows:
\begin{equation}
    \begin{split}
        \mathbb{E}_{q_\phi(z\vert x)}\Bigg[\log\left(\frac{p_\theta(x,z)}{q_\phi(z\vert x)}\right)\Bigg]
        = \int q_\phi(z\vert x)\log \frac{p_\theta(x\vert z)p(z)}{q_\phi(z\vert x)}dz 
        = \int q_\phi(z\vert x)\log \frac{p(z)}{q_\phi(z\vert x)}dz \\+ \int q_\phi(z\vert x)\log p_\theta(x\vert z)dz 
        = - \int q_\phi(z\vert x)\log \frac{q_\phi(z\vert x)}{p(z)}dz + \int q_\phi(z\vert x)\log p_\theta(x\vert z)dz  
        \\= -D_{KL}(q_\phi(z\vert x) \vert \vert p(z)) + \mathbb{E}_{q_\phi(z\vert x)}\left[\log p_\theta(x\vert z)\right]
        = \pazocal{L}(x, \theta, \phi),
    \end{split}
\end{equation}
where $D_{KL}(p\vert \vert q)$ is the Kullback-Leibler divergence between distributions $p$ and $q$, and $\pazocal{L}(x, \theta, \phi)$ is the Evidence Lower BOund (ELBO), whose name comes from Equation \ref{eq:jensen}:
\begin{equation} \label{eq:elbo_1}
    \begin{split}
        \log p_\theta(x) \ge - D_{KL}(q_\phi(z\vert x)\vert \vert p(z)) +  \mathbb{E}_{q_\phi(z\vert x)}\left[\log p_\theta(x\vert z)\right] 
        = \pazocal{L}(x, \theta, \phi),
    \end{split}
\end{equation}
that is, the ELBO is a lower bound for the marginal log-likelihood of the relevant set of points. Thus, by maximizing the ELBO, the log-likelihood of the data is maximized. This would be the optimization problem to solve.

\subsubsection{Implementation}
The ELBO derived from Equation \ref{eq:elbo_1} can be effectively implemented using a DNN-based architecture. However, computing the gradient of the ELBO concerning $\phi$ presents challenges due to the presence of $\phi$ in the expectation term (the second part of the ELBO in Equation \ref{eq:elbo_1}). To address this issue, \cite{kingma} introduced the reparameterization trick. This method involves modifying the latent space sampling process to make it differentiable, enabling the use of gradient-based optimization techniques. Rather than sampling directly from the latent space distribution, VAEs sample $\epsilon$ from a simple distribution, often a standard normal distribution. Subsequently, a deterministic transformation $g_\phi$ is applied to $\epsilon$, producing $z = g_\phi(x, \epsilon)$ where $z \sim q_\phi(z\vert x)$ and $\epsilon \sim p(\epsilon)$. In this case, the ELBO can be estimated as follows.
\begin{equation}\label{eq:elbo_2}
    \begin{split}
         \pazocal{\hat L}(x, \theta, \phi) = \frac{1}{N}\sum_{i=1}^{N} \bigg(- D_{KL}(q_\phi(z\vert x_i)\vert \vert p(z)) + \log p_\theta(x_i\vert g_\phi(x_i, \epsilon_{i}))  \bigg).
    \end{split}
\end{equation}
This modification facilitates the calculation of the ELBO gradient concerning $\theta$ and $\phi$, allowing the application of standard gradient optimization methods. 

\begin{figure}
    \centering
     \includegraphics[width=0.85\textwidth]{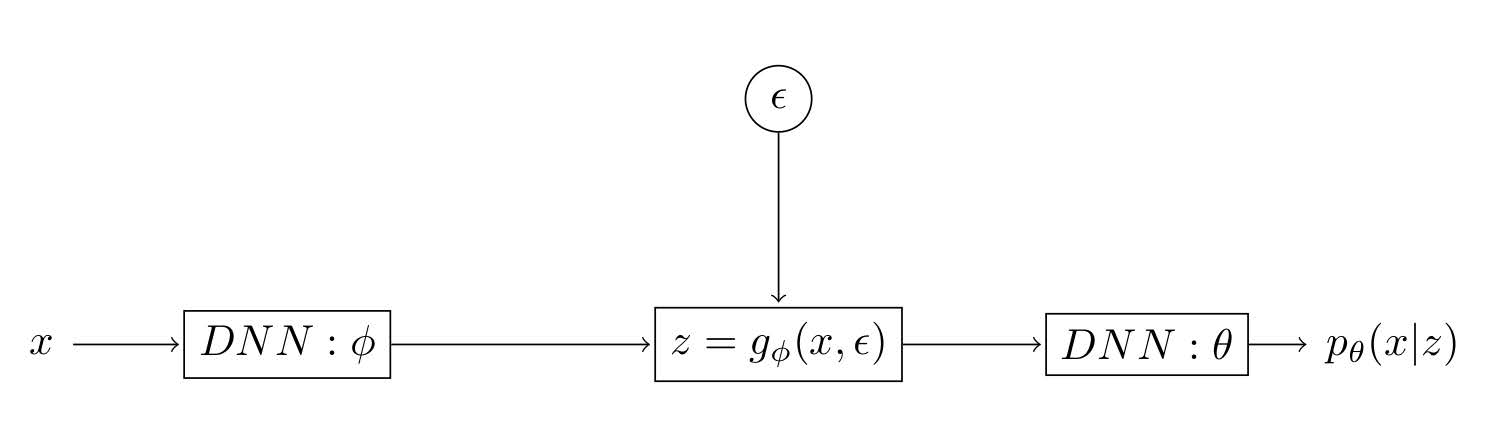}
    \caption{VAE vanilla model implementation using DNNs.}
    \label{fig:dnn_vae}
\end{figure}

Equation \ref{eq:elbo_2} offers a solution using DNNs, with functions parameterized by $\phi$ and $\theta$. Gradients can be conveniently computed using the Backpropagation algorithm, which is automated by various programming libraries. The term VAE derives from the fact that Equation \ref{eq:elbo_2} resembles the architecture of an Autoencoder (AE) \cite{hinton2006reducing}, as illustrated in Figure \ref{fig:dnn_vae}. Notably, the variational distribution $q_\phi$ can be implemented using a DNN with weights $\phi$, taking an input sample $x$ and outputting parameters for the deterministic transformation $g_\phi$. The VAE's latent space comprises the distribution of the latent variable $z$, which is a deterministic transformation $g_\phi$ of the encoder DNN output and random ancillary noise $\epsilon$. A sampled value $z_i$ is drawn from the latent distribution and used to generate an output sample, where another DNN with weights $\theta$ acts as a decoder, taking $z$ as input and providing parameters of the distribution $p_\theta(x\vert z)$ as output.

Two key observations emerge.
\begin{enumerate}
    \item The ELBO losses in Equation \ref{eq:elbo_1} include a regularization term penalizing deviations from the prior in the latent space and a reconstruction error term that enforces similarity between generated samples from the latent space and inputs.
    \item In contrast to standard AEs, VAEs incorporate intermediate sampling, rendering them non-deterministic. This dual sampling process is retained in applications where the distribution of output variables is of interest, facilitating the derivation of input value distribution parameters.
\end{enumerate}

\section{Materials and Methods} \label{s:method}
\begin{figure}
    \centering
     \includegraphics[width=0.5\textwidth]{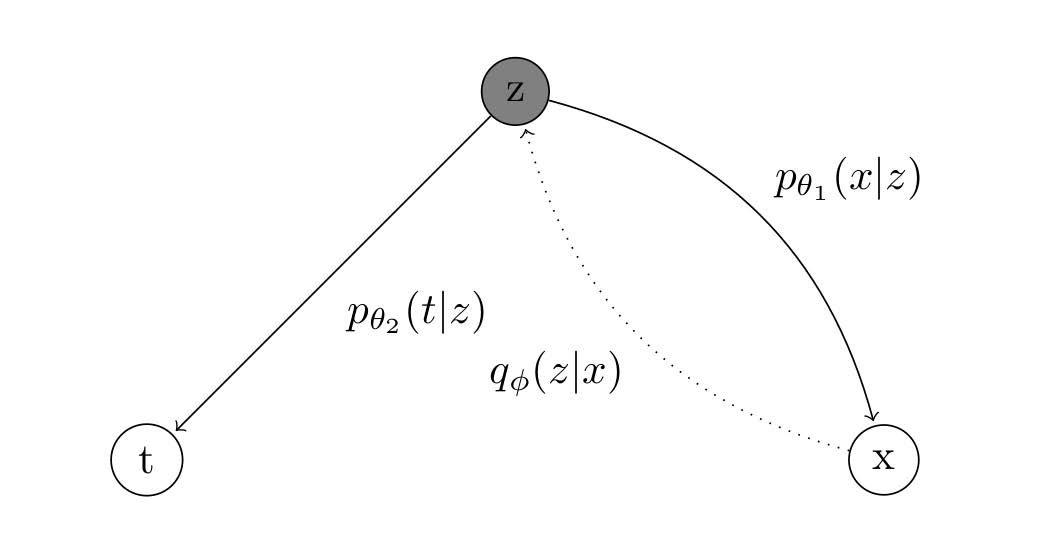}
    \caption{SAVAE Bayesian model. The shadowed circle refers to the latent variable and white circles refer to the observables. Note that the probabilities $p_{\theta_1}(x\vert z)$ and $p_{\theta_2}(t\vert z)$ denote the generative models, and $q_\phi(z|x)$ denotes the variational approximation to the posterior, since the true posterior $p(z|x)$ is unknown.}
    \label{fig:savae}
\end{figure}

The interest lies in using VAEs to obtain the predictive distribution of time-to-event given covariates. The proposed approach, termed Survival Analysis VAE (SAVAE) and depicted in Figure \ref{fig:savae}, extends the Vanilla VAE. SAVAE includes a continuous latent variable $z$, two vectors (an observable covariate vector $x$ and the time-to-event $t$), and generative models $p_{\theta_1}(x\vert z)$ and $p_{\theta_2}(t\vert z)$, assuming conditional independence, which is a characteristic inherent to VAEs and their ability to effectively model the joint distribution of variables. This means that knowing $z$, the components of the vector $x$ and $t$ can be generated independently. To define the predictive distribution based on covariates, a single variational distribution estimates the variational posterior $p(z\vert x)$. While it is possible to include the effect of time ($p(z\vert t, x)$), this approach focuses on using only covariates to obtain the latent space, as the time $t$ can be unknown to predict survival times for test patients and could be censored. SAVAE combines VAEs and survival analysis, offering a flexible framework for modeling complex event data.

\subsection{Goal}
To achieve the main objective, which is to obtain the predictive distribution for the time to event, variational methods will be used in the following way defined in \cite{ranganath}:
\begin{equation}\label{eq:goal}
\begin{split}
    p\left(t^*\vert x^*, \left\{x_i, t_i\right\}^N_{i=1}\right) = \int p \left(t^*\vert z, \left\{x_i, t_i\right\}^N_{i=1}\right) p \left(z\vert x^*, \left\{x_i, t_i\right\}^N_{i=1}\right)dz,
\end{split}
\end{equation}
where $x^*$ represents the covariates of a certain patient, and its survival time distribution $p \left(t^*\vert z, \left\{x_i, t_i\right\}^N_{i=1}\right)$ needs to be estimated. 

\subsection{ELBO derivation}\label{elbo_sa}
Considering our main objective and the use of VAE as the architecture on which we base our approach, the ELBO development seen previously can be extended to apply to our case. SAVAE assumes that the two generative models $p_{\theta_1}(x\vert z)$ and $p_{\theta_2}(t\vert z)$ are conditionally independent. This implies that if $z$ is known, it is possible to generate $x$ or $t$. Furthermore, due to the VAE architecture, it is assumed that each component of the covariate vector $x$ is also conditionally independent given $z$. Therefore, 
\begin{equation}
    p(x, t, z) = p_{\theta_1}(x\vert z)p_{\theta_2}(t\vert z)p(z) = p_\theta(x, t\vert z)p(z).
\end{equation}
It also assumes that the distribution families of $p_{\theta_1}(x\vert z)$ and $p_{\theta_2}(t\vert z)$ are known, but not the parameters $\theta_1$ and $\theta_2$.
Taking into account these assumptions, the ELBO can be computed in a similar way to the case of the Vanilla VAE. First, the conditional likelihood of a set of points $\left\{x_i, t_i\right\}^N_{i=1}$ can be expressed as follows:
\begin{equation} \label{eq:likelihood}
    \begin{split}
        \log p_\theta(x_1, x_2, ..., x_N, t_1, t_2, ..., t_N\vert z) = \sum_{i=1}^{N}\log p_\theta(x_i, t_i\vert z) \\
        = \sum_{i=1}^{N}\left(\log p_{\theta_2}(t_i\vert z) + \sum_{l=1}^{L} \log p_{\theta_{1}}(x_{i}^{l}\vert z)\right),
    \end{split}
\end{equation}
where the expected conditional likelihood can be expressed as:
\begin{equation}
    \begin{split}
        \mathbb{E}_z\left[p_\theta(x, t\vert z)\right] 
        = \int p_\theta(x, t\vert z)p(z)dz 
        = \int\frac{p_\theta(x, t, z)}{p(z)}p(z)dz 
        \\=
        \int p_\theta(x, t, z)dz
        =p_\theta(x, t) = \int p_\theta(x, t, z)\frac{q_\phi(z\vert x)}{q_\phi(z\vert x)}dz
        \\= \mathbb{E}_{q_\phi(z\vert x)}\left[\frac{p_\theta(x,t,z)}{q_\phi(z\vert x)}\right].
    \end{split}
\end{equation}
As the interest lies in computing the log-likelihood:
\begin{equation}
    \begin{split}
        \log p_\theta(x, t) = \log\left[\mathbb{E}_{q_\phi(z\vert x)}\left[\frac{p_\theta(x,t,z)}{q_\phi(z\vert x}\right]\right]
       \\ \ge \mathbb{E}_{q_\phi(z\vert x)}\left[\log\frac{p_\theta(x,t,z)}{q_\phi(z\vert x)}\right],
    \end{split}
\end{equation}
where the inequality comes from applying Jensen's inequality. Then, this could be rearranged as:
\begin{equation}
    \begin{split}
        \mathbb{E}_{q_\phi(z\vert x)}\left[\log\left(\frac{p_\theta(x,t,z)}{q_\phi(z\vert x)}\right)\right]
        = \int q_\phi(z\vert x)\log \frac{p_{\theta_1}(x\vert z)p_{\theta_2}(t\vert z)p(z)}{q_\phi(z\vert x)}dz \\
        = - \int q_\phi(z\vert x)\log \frac{q_\phi(z\vert x)}{p(z)}dz
        + \int q_\phi(z\vert x)\left(\log p_{\theta_1}(x\vert z)+\log p_{\theta_2}(t\vert z)\right)dz \\
        = -D_{KL}(q_\phi(z\vert x) \vert \vert p(z)) + \mathbb{E}_{q_\phi(z\vert x)}\left[\log p_{\theta_1}(x\vert z) + \log p_{\theta_2}(t\vert z)\right] \\
        = \pazocal{L}(x, \theta_1, \theta_2, \phi).
    \end{split}
\end{equation}
After computing this ELBO, it can be seen that it is similar to the Vanilla VAE's one (Equation \ref{eq:elbo_2}). The only difference lies in the reconstruction term, which is expressed differently in order to explicitly distinguish between the covariates and the time-to-event. By using Equation \ref{eq:likelihood} and the reparameterization trick, the ELBO estimator is obtained, explicitly accounting for each dimension of the covariates vector:
\begin{equation}\label{eq:elbo_3}
\begin{split}
    \pazocal{\hat L}(x, \theta_1, \theta_2, \phi)
    \\= \frac{1}{N} \sum_{i=1}^{N}\Bigg(- D_{KL}(q_\phi(z\vert x_i)\vert \vert p(z)) + \log p_{\theta_2}(t_i\vert g_\phi(x_i, \epsilon_{i})) 
    + \sum_{l=1}^{L}\log p_{\theta_{1}}(x_{i}^{l}\vert g_\phi(x_i, \epsilon_{i}))\Bigg).
\end{split}
\end{equation}
\begin{figure*}
    \includegraphics[width=\textwidth]{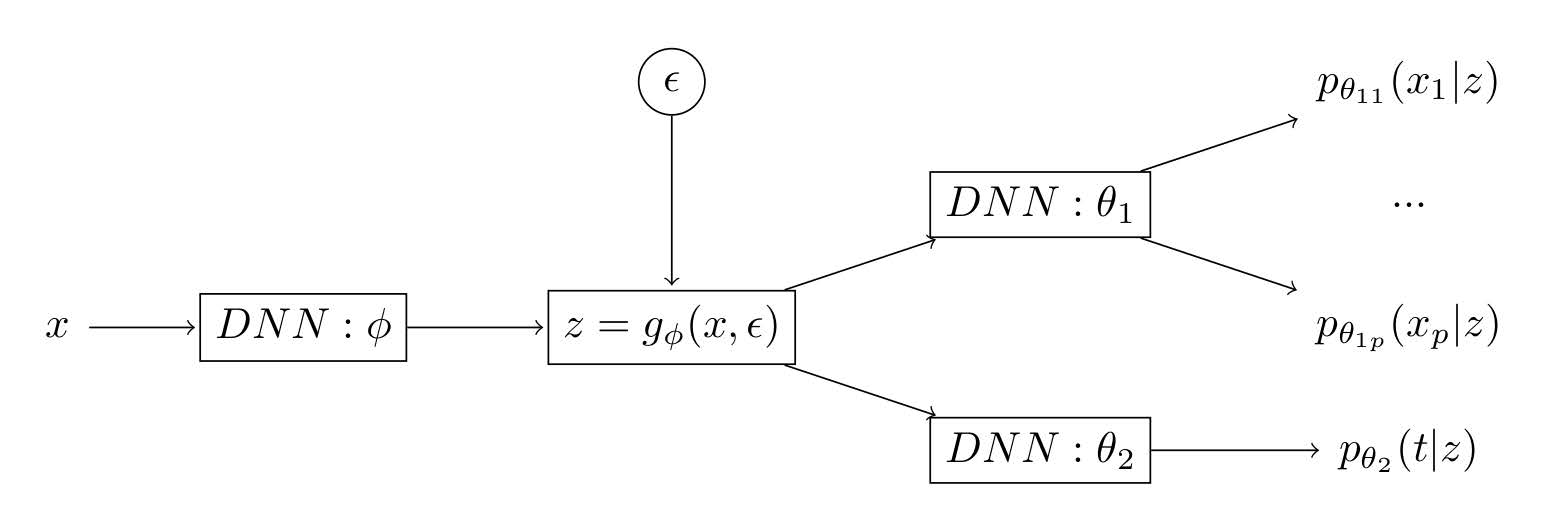}
    \caption{SAVAE implementation using DNNs. One of them acts as an encoder which has the covariates vector as input. The other two act as decoders, one for the covariates and the other one for the time.}
    \label{fig:savae_dnns}
\end{figure*}

In terms of implementation, three DNNs have been used, as specified in Figure \ref{fig:savae_dnns}. Note that the decoder DNNs output the parameters of each distribution. 

\subsubsection{Divergence computation}
SAVAE assumes that $q_\phi(z\vert x)$ follows a multidimensional Gaussian distribution defined by a vector of means $\mu$, where each element is $\mu_j$ and by a diagonal covariance matrix $\textbf{C}$, where the main diagonal consists of variances $\sigma^2_j$. According to \cite{kingma}, it can be stated that:
\begin{equation}
    - D_{KL}(q_\phi(z\vert x)\vert \vert p(z)) = \frac{1}{2} \sum_{j=1}^{J} (1 + \log(\sigma_j^2) - \mu_j^2 -\sigma_j^2),
\end{equation}
where $J$ is the dimension of the latent space $z$. This means that the Kullback-Leibler divergence from the ELBO Equation \ref{eq:elbo_3} can be calculated analytically.

\subsubsection{Time modeling}\label{time_mod}
One significant challenge in handling survival data is the issue of censorship, which occurs when a patient has not yet experienced the event of interest. In such cases, the true survival time remains unknown, resulting in partial or incomplete observations. Consequently, SA models must employ specific techniques capable of accommodating censored observations along with uncensored ones to reliably estimate relevant parameters.

In our case, to account for censoring in survival data, we start from the time $t$ reconstruction term from Equation \ref{eq:elbo_3} for a single patient: 
\begin{equation}
    \pazocal{\hat{L}}_{time}(x_i, \theta_2, \phi) = \log p_{\theta_2}(t_i\vert g_\phi(x_i, \epsilon_{i})).
\end{equation}
Taking into account the censoring indicator $d_i$:
\begin{equation}
d_i = 
        \begin{cases}
        0 \;\;\;\text{if censored}\\
        1 \;\;\;\text{if event experienced}
        \end{cases},
\end{equation}
we could just use the information given by uncensored patients. However, we would waste information, since we know that the censored patients have not experienced the event until time $t_i$. Hence, considering Equation \ref{eq:sa} and following \cite{liverani}, we model the time pdf as:
\begin{equation} \label{eq:time_pdf}
    p_{\theta_2}(t_i\vert g_\phi(x_i, \epsilon_{i})) = h(t_i \vert g_\phi(x_i, \epsilon_{i}))^{d_i}S(t_i \vert g_\phi(x_i, \epsilon_{i})).
\end{equation}
Therefore, the hazard function term is only taken into account when the event has been experienced, that is, when the data are not censored. This way, SAVAE incorporates information from censored observations, providing consistent parameter estimates.

Regarding the distribution chosen for the time event, we have followed several publications such as \cite{ranganath_2016} where the Weibull distribution model is used. This distribution is two-parameter, with positive support, that is, $p(t) = 0, \forall t < 0$. The two scalar parameters of the distribution are $\lambda$ and $\alpha$, where $\lambda > 0$ controls the scale and $\alpha > 0$ controls the shape as follows:
\begin{equation}
    \begin{cases}
     p(t; \alpha, \lambda) = \frac{\alpha}{\lambda} \left( \frac{t}{\lambda} \right)^{\alpha - 1} \exp{ \left(-\left(\frac{t}{\lambda}\right)^\alpha\right)}\\
     S(t; \alpha, \lambda) = \frac{}{} \exp{\left(-\left(\frac{t}{\lambda}\right)^\alpha\right)}\\
     h(t; \alpha, \lambda) = \frac{p(t; \alpha, \lambda)}{S(t; \alpha, \lambda)} = \frac{\alpha}{\lambda}\left(\frac{t}{\lambda}\right)^{\alpha-1}
     \end{cases}.
\end{equation}
Although the Weibull distribution is our primary choice for modeling time-to-event data in SAVAE, it is crucial to highlight that other distributions are feasible, as long as their hazard functions and CDFs can be analytically calculated. This versatility distinguishes SAVAE from other models. For example, the exponential distribution, a special case of Weibull with $\alpha = 1$, can represent constant hazard functions. Integrating alternative distributions, such as the exponential, into SAVAE is straightforward and only requires adjusting the terms in Equation \ref{eq:time_pdf}. The ability of SAVAE to predict the distribution parameters for each patient facilitates the calculation of various statistics, such as means, medians and percentiles, providing flexibility beyond the models customized to a single distribution.

\subsubsection{Marginal log-likelihood computation}\label{marginal_ll}
Assigning distribution models to patient covariates in the reconstruction term is essential in SAVAE. This choice enables control over the resulting output variable distribution, but it also implies that the model approximates the chosen distribution even if the actual distribution differs. The third component of the ELBO (\ref{eq:elbo_3}) depends on the log-likelihood of the data, which for some representative distributions is: 

\begin{itemize}
    \item \textbf{Gaussian distribution: }Suitable for real-numbered variables ($x_{i}^{l} \in (-\infty, +\infty)$), it has parameters $\mu \in (-\infty, +\infty)$ and $\sigma \in (0, +\infty)$, known for its symmetric nature. Its log-likelihood function is:
    \begin{equation}
        \begin{split}
            \log(p(x_{i}^{l};\mu, \sigma)) = -\log(\sigma \sqrt{2\pi}) - \frac{1}{2} \left(\frac{x_{i}^{l}-\mu}{\sigma}\right)^2.
        \end{split}
    \end{equation}
    \item \textbf{Bernoulli distribution: }Applied to binary variables ($x_{i}^{l} \in \{0,1\}$), it has a single parameter $\beta \in [0,1]$, representing the probability of $x_{i}^{l}=1$.Its log-likelihood function is:
    \begin{equation}
        \log(p(x_{i}^{l};\beta)) = x_{i}^{l}\log(\beta) + (1-x_{i}^{l})\log(1-\beta).
    \end{equation}
    \item \textbf{Categorical distribution: }Models discrete variables with $K$ possible values. We can think of $x_{i}^{l}$ as a categorical scalar random variable with $K$ different values. Each possible outcome is assigned a probability $\theta_k$ (note that $\sum_{k=1}^K \theta_k = 1$). The log-likelihood function can be computed based on the Probability Mass Function (PMF) following the expression:
    \begin{equation}
        \log(p(x_{i}^{l} \vert \theta_1, \theta_2, ..., \theta_k)) = \log\left(\prod_{k=1}^K \theta_k^{\mathbb{I}(x_{i}^{l}=k)}\right),
    \end{equation}
    where the indicator function means:
    \begin{equation}
        \mathbb{I}(x_{i}^{l}=k) =   
        \begin{cases}
            1  \quad x_{i}^{l} = k\\
            0  \quad x_{i}^{l} \neq k
     \end{cases}.
    \end{equation}
\end{itemize}

Recall that other desired distributions can be implemented in SAVAE, as long as their log-likelihood is differentiable.

\section{Results and Discussion}
Once SAVAE has been defined, the next step is to proceed with the experimental validation. First, the data used as input to the model will be discussed, followed by the experimental setup (network architecture and training process). Finally, SAVAE's performance evaluation will be analyzed. The code can be found in \url{https://github.com/Patricia-A-Apellaniz/savae}.

\subsection{Survival data}
\renewcommand{\arraystretch}{1.5}
\begin{table*}[t]
\resizebox{\textwidth}{!}{
\centering
\begin{tabular}{cccccc}
\hline
\textbf{Dataset} & \textbf{\# Samples}& \textbf{\# Censored} & \textbf{\# Covariates} & \textbf{Event Time (mean, (min - max))} & \textbf{Censoring Time (mean, (min - max))} \\
\hline
WHAS       & 1638  &           948 (57.88\%)              &     5                  &         1045.42 (1 - 1999) days             &         1298.92 (371 - 1999) days         \\
SUPPORT     & 9104    &         2904 (31.89\%)              &            14           &          478.45 (3 - 2029) days           &              1060.22 (344 - 2029) days          \\ 
GBSG       & 1546       &        965 (43.23\%)               &          7             &         44.49 (0.26 - 87.36) months           &         65.15 (0.26 - 87.36) months              \\    
FLCHAIN       & 6524       &         4562 (69.92\%)              &           8            &         3647.5 (0 - 5166) days           &         4296.74 (1 - 5166) days               \\ 
NWTCO       & 4028       &         3457 (85.82\%)              &           6            &         2276.68 (4 - 6209) days           &         2588.23 (4 - 6209) days               \\ 
METABRIC       & 1980       &         854 (56.18\%)              &           21            &         2944.81 (3 - 9193) days           &         3424.81 (21 - 9193) days               \\ 
PBC        & 418      &         257 (61.48\%)              &           17            &         63.93 (1.37 - 159.8) months           &         75.22 (17.77 - 159.83) months               \\ 
STD       & 877       &         530 (60.43\%)              &           21            &         369 (1 - 1519) days           &         420 (1 - 1519) days               \\ 
PNEUMON     & 3470         &         3397 (97.9\%)              &           13            &         9.84 (0.5 - 12) months           &         9.98 (0.5 - 12) months              \\ 
\hline
\end{tabular}}
\caption{Data information from datasets used to train SAVAE model. We have analyzed nine different disease datasets with different proportions of samples, censored data, and varying survival times. Additionally, each contains different patient information, be it genomic, clinical, or demographic data.}
\label{table:datasets}
\end{table*}

In SA datasets, each patient contributes information about whether events of interest occurred during a study period, categorizing them as censored or uncensored, along with their respective follow-up times. To evaluate SAVAE, we trained it in nine diverse disease datasets, including WHAS, SUPPORT, GBSG, FLCHAIN, NWTCO, METABRIC, PBC, STD, and PNEUMON. We followed pre-processing procedures similar to state-of-the-art models, ensuring a fair evaluation on established benchmarks in SA.

The Worcester Heart Attack Study (WHAS) \cite{whas} focuses on patients with acute myocardial infarction (AMI), providing clinical and demographic data. The Study to Understand Prognoses Outcomes and Risks of Treatment (SUPPORT) \cite{support} investigates seriously ill hospitalized adults and includes information on demographics, comorbidities, and physiological measurements. The Rotterdam \& German Breast Cancer Study Group (GBSG) \cite{rott, gbsg} combines data from node-positive breast cancer patients and a chemotherapy trial. The FLCHAIN \cite{flchain} dataset studies the relationship between mortality and serum immunoglobulin Free Light Chains, which are important in hematological disorders. NWTCO \cite{nwtco} studies Wilms tumor in children, Molecular Taxonomy of Breast Cancer International Consortium (METABRIC) \cite{metabric} explores breast cancer, PBC focuses on Primary Biliary Cholangitis, STD deals with sexually transmitted diseases, and PNEUMON examines infant pneumonia.

Table \ref{table:datasets} offers a more comprehensive view of the temporal aspects and occurrences of events within the various datasets considered. It becomes evident that a deliberate selection of various disease datasets has been made, each characterized by distinct types and quantities of information. Significantly, the evaluation of the model has been carried out systematically in datasets that show varying proportions of censored samples and differing time-to-event ranges. This strategic approach aims to provide a broader perspective on how the model might perform when applied to other real-world datasets.

\subsection{Performance metrics}
Recalling from Section \nameref{sa}, each dataset is described by $D = (x_i, t_i, d_i)^N_{i=1}$ triplets, where $x_i = x_{i}^{1}, ..., x_{i}^{L}$ is an \textit{L}-dimensional vector of covariates, $t_i$ is the time to event and $d_i \in \{0,1\}$ is the censoring indicator.

When evaluating an SA model, the literature shows that the most commonly used metric is the C-index, which is the generalization of the ROC curve for all data. It is a measure of the rank correlation between predicted risk and observed times. The concept arises from the intuition that a higher risk of an event occurring has a complete relation with a short time to the event. Therefore, a high number of correlating pairs, i.e. pairs of samples that meet this expectation, is decisive to say that the model has good predictive quality.

In this case, the time-dependent C-index described in \cite{antolini} will be used since the original one \cite{harrel} cannot reflect the possible changes in risk over time being only computed at the initial time of observation. This C-index is defined as follows:
\begin{equation}
    \begin{split}
            C_{index} = P\Big( \hat F(t\vert x_i) > \hat F(t\vert x_j)\vert d_i = 1, t_i <t_j, t_i \leq t \Big),
    \end{split}
\end{equation}
where $\hat F(t\vert x_i)$ is the CDF estimated by the model at the time $t$ given a set of covariates $x_i$. The probability is estimated by comparing the relative risks pairwise, as already mentioned. 

Based on the prediction index defined in \cite{brier}, \cite{graf} proposed the second evaluation metric that has been used in this analysis: Brier Score (BS). It is essentially a square prediction error based on the Inverse Probability of Censoring Weighting (IPCW) \cite{ipcw}, a technique designed to recreate an unbiased scenario compensating for censored samples by giving more weight to samples with similar features that are not censored. So, given a time $t$ the BS can be calculated as follows, with $G(\cdot)$ being the survival function corresponding to censoring ($1/G(t)$ is the IPCW):
\begin{equation}
    \begin{split}
    BS(t) = \frac{1}{N} \sum_{i=1}^N \Bigg[ \frac{ (S(t\vert x_i))^2}{G(t_i)} \cdot \mathbb{I}(t_i < t, d_i=1) \\+ \frac{(1 - S(t\vert x_i))^2}{G(t)} \cdot \mathbb{I}(t_i \geq t) \Bigg].
    \end{split}
\end{equation}
Since the C-index does not take into account the actual values of the predicted risk scores, BS can be used to assess calibration, i.e., if a model predicts a 10\% risk of experiencing an event at time \textit{t}, the observed frequency in the data should match this percentage for a well-calibrated model. On the other hand, it is also a measure of discrimination: whether a model can predict risk scores that allow us to correctly determine the order of events. 

In this case, the evaluation is made using the integral form of BS since it does not depend on the selection of a specific time $t$:
\begin{equation}
    IBS(t_{max}) = \frac{1}{t_{max}}\int_0^{t_{max}}BS(t)dt.
\end{equation}

To statistically assess the performance of each model based on the C-index globally, we propose the Mean Reciprocal Rank (MRR) as the third metric. It measures the effectiveness of a prediction by considering the rank of the first relevant C-index within a list composed of the C-indices obtained from each model. Formally, the Reciprocal Rank (RR) for a set of results for each model is the inverse of the position of the first pertinent result. For example, if the first relevant result is in position 1, its RR is 1; if it is in position 2, the RR is 0.5; if it is in position 3, the RR is approximately 0.33, and so on. Thus, the MRR is the average of the RRs for a set of models:
\begin{equation}
    MRR = \frac{1}{Q}\sum_{i=1}^{Q}\frac{1}{rank_i},
\end{equation}

where $Q$ is the total number of models that are being compared, and $rank_i$ is the position of the first relevant C-index for the $i-th$ model. Higher MRR values indicate that relevant results tend to appear higher in the list.

Finally, to add more statistical information on the performance of the models, we performed hypothesis testing to compare the mean C-index and IBS values of our model with those of the state-of-the-art models in multiple folds, since we are using a five-fold cross-validation method. Specifically, we formulated a null hypothesis that assumes that the mean performance metrics of the state-of-the-art models are greater than our model's mean performance metrics. To assess the validity of this null hypothesis, we used $p$-values as a statistical measure. We established a significance threshold of 0.05, a common practice in hypothesis testing. When the obtained $p$-value for each case fell below this threshold, we rejected the null hypothesis. In practical terms, this indicated that our model exhibited superior performance compared to the other models. On the contrary, if the $p$-value exceeded 0.05, we concluded that there were no statistically significant differences between our model and the others. It is important to note that this approach considered variations in results across different folds, providing a more comprehensive assessment of model performance beyond just the average results.

\subsection{Experimental setting}\label{exp_setting}
To begin with, the implementation of SAVAE was executed using the PyTorch framework \cite{paszke2019pytorch}. As defined in Section \ref{elbo_sa}, three different DNNs were trained, consisting of one encoder and two decoders. These decoders were designed to infer covariates and time parameters, respectively. The Gaussian encoder exhibits a straightforward architecture, characterized by a single hidden linear layer featuring a Rectified Linear Unit (ReLU) activation function and an output linear layer with hyperbolic tangent activation. The input to this encoder consists of the covariate vectors from the training dataset, while the output generates a Gaussian latent space. The dimensionality of this latent space has been fixed to 5. The generated latent space serves as input for both decoders, each featuring two linear layers. The first layer employs a ReLU activation function and incorporates a dropout rate of 20\%. However, the final layer of the decoders employs different activation functions based on the specified distribution, thereby tailoring the output to the parameters of the respective covariate distribution. Furthermore, the number of neurons in each hidden layer was also fixed at 50. The training process involved 3000 epochs with a batch size of 64 samples while incorporating an Early Stop mechanism in the event of an insufficient reduction in validation loss.

To evaluate the results while ensuring their robustness against data partitioning, we used a five-fold cross-validation technique. This method was applied not only to our model but also to the state-of-the-art models used for performance comparison and result evaluation, including Cox-PH, DeepHit, and DeepSurv. Moreover, due to the inherent sensitivity of VAE architectures to initial conditions, we conducted training using up to 10 different random seeds. Subsequently, the C-index was averaged among the three best performing seeds. We consider that the average performance of three seeds provides a representative and sufficient evaluation. Lastly, note that the three state-of-the-art models have been implemented using the Pycox package \cite{pycox}, as well as the different metrics used for validation, C-index and IBS. The MRR has been calculated manually, while the $p$-value has been obtained using the SciPy \cite{scipy} package.

\subsection{Results}
In this section, we present a comprehensive assessment of the performance of our proposed model, SAVAE, compared to three well-established state-of-the-art models. Cox-PH, DeepSurv, and DeepHit. Across multiple datasets that encompass a diverse range of medical and clinical scenarios, we conducted extensive experiments to assess the performance of these models. The key focus was on evaluating their ability to predict survival outcomes, considering censored and uncensored data points. 

As the initial set of results, our focus is on comparing the performance and results in terms of the C-index. Table \ref{table:c_index_results} provides a comprehensive view of how our model is completely comparable to the state-of-the-art models in terms of the average C-index. Additionally, note that all intervals for the minimum and maximum values across various folds overlap, indicating consistent performance across different data subsets. The results displayed in the table reveal that our model consistently achieves a higher MRR compared to others across multiple datasets, showcasing its superiority in many cases regarding the average C-index. However, it is essential to acknowledge that the C-index results among the different models are generally similar, highlighting the competitiveness of our model within the field. Furthermore, it is important to note that the broad intervals are primarily attributed to the limited sample sizes commonly found in medical databases, a characteristic that poses challenges when assessing model performance. To address this issue, we employed cross-validation as previously mentioned, ensuring that our model's performance is robust and reliable. In summary, while our model demonstrates its strength by outperforming other models in terms of MRR and achieving competitive average C-index scores, the overall similarity in C-index results underscores its robustness and suitability for various medical datasets.
\newenvironment{sfsmtabbing}
  {\sffamily\scriptsize\tabbing}
  {\endtabbing\par}

\setlength{\tabcolsep}{8pt}
\renewcommand{\arraystretch}{1.5}
\begin{table*}
\resizebox{\textwidth}{!}{
\begin{tabular}{lcccccccc}
 \hline
\multicolumn{1}{l}{\multirow{2}{*}{Dataset}} & \multicolumn{2}{c}{COXPH}   & \multicolumn{2}{c}{DEEPSURV} & \multicolumn{2}{c}{DEEPHIT} & \multicolumn{2}{c}{SAVAE}  \\
\multicolumn{1}{c}{}                         & Avg. C-index & (min, max)      & Avg. C-index  & (min, max)   & Avg. C-index & (min, max)      & Avg. C-index & (min, max) \\
 \hline
WHAS  & 0.74     & (0.66, 0.81) & 0.78   & (0.57, 0.88) & \textbf{0.89}  & (0.82, 0.95) & 0.74 & (0.67, 0.80) \\ 
SUPPORT  & 0.58     & (0.39, 0.78) & 0.57   & (0.37, 0.82) & 0.55  & (0.37, 0.73) & \textbf{0.61} & (0.40, 0.86) \\
GBSG  & 0.66     & (0.61, 0.71) & \textbf{0.67}   & (0.58, 0.73) & 0.66  & (0.58, 0.72) & \textbf{0.67} & (0.62, 0.72) \\
FLCHAIN  & 0.69     & (0.50, 0.80) & 0.67   & (0.55, 0.80) & 0.78  & (0.73, 0.82) & \textbf{0.79} & (0.75, 0.83) \\
NWTCO  & 0.71     & (0.64, 0.79) & 0.70   & (0.60, 0.79) & \textbf{0.72}  & (0.66, 0.78) & 0.71 & (0.63, 0.79) \\
METABRIC  & 0.59     & (0.52, 0.68) & \textbf{0.61}   & (0.52, 0.69) & 0.56  & (0.46, 0.64) & \textbf{0.61} & (0.53, 0.70) \\
PBC  & \textbf{0.81}     & (0.64, 0.94) & 0.80   & (0.65, 0.92) & 0.80  & (0.62, 0.93) & \textbf{0.81} & (0.62, 0.95) \\
STD  & \textbf{0.60}     & (0.47, 0.72) & \textbf{0.60}   & (0.49, 0.71) & 0.59  & (0.50, 0.68) & 0.59 & (0.46, 0.71) \\
PNEUMON  & 0.62     & (0.54, 0.70) & 0.65   & (0.49, 0.80) & \textbf{0.67}  & (0.57, 0.77) & 0.65 & (0.53, 0.77) \\ \hline
MRR           & \multicolumn{2}{c}{0.56} & \multicolumn{2}{c}{0.60} & \multicolumn{2}{c}{0.62} & \multicolumn{2}{c}{\textbf{0.76}} \\ \hline
\end{tabular}}
\begin{sfsmtabbing}
\hspace{125pt}\textbf{Bold} highlights the best mean. For C-index and MRR, higher is better  \=
\end{sfsmtabbing}
\caption{C-index average results across different folds for each state-of-the-art model. Average C-index results across the three best seeds for each fold in SAVAE performance. MRR values are given to rank each model attending only to the mean value.}
\label{table:c_index_results}
\end{table*}
\setlength{\tabcolsep}{8pt}
\renewcommand{\arraystretch}{1.5}
\begin{table*}
\resizebox{\textwidth}{!}{
\begin{tabular}{lccccccccc}
 \hline 
Model & WHAS   & SUPPORT & GBSG & FLCHAIN & NWTCO &  METABRIC & PBC & STD  & PNEUMON\\
 \hline
COXPH  & 0.579    & 0.058  & \textbf{0.0} & \textbf{0.0} & 0.268 & \textbf{0.003} & 0.45 & 0.887 & \textbf{0.003}  \\ 
DEEPSURV    &  1.0 & \textbf{0.02} & 0.149  & \textbf{0.0} & 0.135 & 0.549 & 0.28 & 0.927 & 0.382 \\ 
DEEPHIT   &  1.0 & \textbf{0.0}& \textbf{0.0}  & \textbf{0.01} & 0.644 & \textbf{0.0} & 0.228 & 0.727 & 0.935 \\ 
\hline
\end{tabular}}
\begin{sfsmtabbing}
\hspace{55pt}\textbf{Bold} Implies a $p$-value below our threshold, 0.05. This means that SAVAE is significantly better than the other models.  \=
\end{sfsmtabbing}
\caption{$p$-values obtained to determine whether the mean of SAVAE is greater than the state-of-the-art folds C-indexes.}
\label{table:c_index_p_values}
\end{table*}

In our validation process, we performed a statistical analysis using $p$-values to determine whether our model exhibited superior performance in terms of the C-index. To carry out this analysis, we compared the average C-index of our model with the mean C-index values obtained from multiple folds for each of the state-of-the-art models. The objective was to determine whether the performance of our model was statistically better than the alternative models. We established a significance threshold of 0.05, a common practice in hypothesis testing. Our findings in Table \ref{table:c_index_p_values} reveal several instances in which our model outperformed the state-of-the-art models, as evidenced by $p$-values below the 0.05 threshold. These results highlight the effectiveness and competitiveness of our proposed approach. This comprehensive analysis, which considers the diverse C-index values in multiple folds, provides a robust evaluation of the performance of the model, extending beyond simple average comparisons.

Our validation through IBS values (Tables \ref{table:ibs_results} and \ref{table:ibs_p_values}) yielded conclusions that closely parallel those derived from the C-index analysis. Overall, it is important to note that our model's IBS results align closely with those of the state-of-the-art models, demonstrating comparable performance. However, our proposed model consistently demonstrated competitiveness and emerged as the top performer in the various datasets used in our study. This convergence of results across different evaluation metrics reinforces the robustness and effectiveness of our novel approach. While our model maintains a competitive edge within the context of the state-of-the-art models, further solidifying its potential and utility in the field of SA, it also stands out as a top-performing solution.

\setlength{\tabcolsep}{8pt}
\renewcommand{\arraystretch}{1.5}
\begin{table*}
\resizebox{\textwidth}{!}{
\begin{tabular}{lcccccccc}
 \hline
\multicolumn{1}{l}{\multirow{2}{*}{Dataset}} & \multicolumn{2}{c}{COXPH}   & \multicolumn{2}{c}{DEEPSURV} & \multicolumn{2}{c}{DEEPHIT} & \multicolumn{2}{c}{SAVAE}  \\
\multicolumn{1}{c}{}                         & Avg. IBS & (min, max)      & Avg. IBS  & (min, max)   & Avg. IBS & (min, max)      & Avg. IBS & (min, max) \\
 \hline
WHAS  & 0.171     & (0.109, 0.279) & 0.134   & (0.067, 0.260) & \textbf{0.120}  & (0.067, 0.175) & 0.159 & (0.114, 0.205) \\ 
SUPPORT  & 0.208     & (0.074, 0.374) & \textbf{0.205}   & (0.057, 0.363) & 0.219  & (0.086, 0.370) & 0.208 & (0.063, 0.385) \\ 
GBSG  & 0.182     & (0.142, 0.223) & 0.179   & (0.137, 0.228) & 0.208  & (0.168, 0.248) & \textbf{0.179} & (0.139, 0.222) \\ 
FLCHAIN  & 0.137     & (0.089, 0.185) & 0.142   & (0.088, 0.186) & 0.121  & (0.098, 0.145) &\textbf{ 0.102} & (0.078, 0.124) \\ 
NWTCO  & \textbf{0.107}     & (0.080, 0.138) & 0.109   & (0.082, 0.149) & 0.111 & (0.083, 0.147) & 0.127 & (0.101, 0.152) \\ 
METABRIC  & 0.186     & (0.137, 0.233) & 0.191   & (0.143, 0.244) & 0.214  & (0.153, 0.275) & \textbf{0.180} & (0.127, 0.236) \\ 
PBC  & 0.147     & (0.043, 0.281) & 0.146   & (0.046, 0.268) & 0.195  & (0.087, 0.340) & \textbf{0.138} & (0.034, 0.267) \\ 
STD  & 0.210    & (0.121, 0.302) & 0.212   & (0.123, 0.305) & 0.224  & (0.142, 0.315) & \textbf{0.209} & (0.121, 0.307) \\ 
PNEUMON  & \textbf{0.016}     & (0.004, 0.031) & 0.017   & (0.004, 0.034) &\textbf{ 0.016}  & (0.004, 0.031) & 0.021 & (0.007, 0.037) \\ \hline
MRR           & \multicolumn{2}{c}{0.55} & \multicolumn{2}{c}{0.55} & \multicolumn{2}{c}{0.47} & \multicolumn{2}{c}{\textbf{0.71}} \\ \hline

\end{tabular}}
\begin{sfsmtabbing}
\hspace{125pt}\textbf{Bold} highlights the best mean. For IBS lower is better and for MRR, higher is better \=
\end{sfsmtabbing}
\caption{IBS average results across different folds for each state-of-the-art model. Average IBS results in the three best seeds for each fold in SAVAE performance. MRR values are given to rank each model.}
\label{table:ibs_results}
\end{table*}

\setlength{\tabcolsep}{8pt}
\renewcommand{\arraystretch}{1.5}
\begin{table*}
\resizebox{\textwidth}{!}{
\begin{tabular}{lccccccccc}
 \hline
Model & WHAS   & SUPPORT & GBSG & FLCHAIN & NWTCO &  METABRIC & PBC & STD  & PNEUMON\\
 \hline 
COXPH  & 1    & 0.47  & 0.998 & 1 & \textbf{0.0} & 0.995 & 0.888 & 0.575 & \textbf{0.0}  \\ 
DEEPSURV    &  \textbf{0.0} & 0.341 & 1  & \textbf{0.0} & 1.0 & 0.549 & 0.868 & 0.746 & \textbf{0.0} \\
DEEPHIT   &  \textbf{0.0} & 0.950 & 1  & 1 & \textbf{0.0} & 1 & 1.0 & 0.995 & \textbf{0.0} \\ 
\hline
\end{tabular}}
\begin{sfsmtabbing}
\hspace{55pt}\textbf{Bold} Implies a $p$-value below our threshold, 0.05. This means that SAVAE is significantly better than the other models.  \=
\end{sfsmtabbing}
\caption{$p$-values obtained to determine whether the mean of SAVAE is greater than the state-of-the-art folds C-indexes.}
\label{table:ibs_p_values}
\end{table*}

\section{Conclusions}
In this paper, we have successfully described an SA model (SAVAE), which stands out for its ability to avoid assumptions that can limit performance in real-world scenarios. It is a model based on VAEs in charge of estimating continuous or discrete survival times, first, modeling complex non-linear relations among covariates due to the use of highly expressive DNNs, and second, taking advantage of a combination of loss functions that capture the censoring inherent to survival data. Our model demonstrates efficiency compared to various state-of-the-art models, namely Cox-PH, DeepSurv, and DeepHit, because of its freedom from assumptions related to linearity and proportional hazards. In contrast to DeepHit, which directly learns the C-Index metric, we train using standard likelihood techniques. Note that this means that our approach is more flexible, as it allows using many different distributions to model the data, and the performance is competitive, as it performs well in C-Index and IBS.

Furthermore, the adaptability of our model is a notable strength. While we have assumed specific distributions for both survival times and covariates in our experiments, SAVAE's versatility extends to accommodating any other parametric distribution, as long as their CDF and hazard function are differentiable, making it a scalable tool. Notably, our model can efficiently handle censoring to mitigate bias, introducing a novel improvement in results.

This work raises several attractive lines for the future. An additional advantage lies in our model's architecture, where time and covariates are reconstructed from latent space information. This feature opens opportunities for its utility to be expanded to various tasks that have been developed using VAEs, including clustering \cite{vae_clustering}, imputation of missing data \cite{vae_imputation}, and data augmentation \cite{vae_augmentation} by the generation of synthetic patients. Thus, this tool has great potential and can be exploited in future work to have different functionalities even in the world of Federated Learning \cite{vae_fl} \cite{vae_fl_2}.

In summary, SAVAE emerges as a versatile and robust model for SA, surpassing state-of-the-art methods while offering extensibility to a broader range of healthcare applications. It presents a compelling solution for healthcare professionals looking for enhanced performance and adaptability in SA tasks.

\section*{Acknowledgments}
This research was supported by GenoMed4All project. GenoMed4All has received funding from the European Union’s Horizon 2020 research and innovation programme under grant agreement No 101017549. The authors declare that they have no known competing financial interests or personal relationships that could have appeared to influence the work reported in this paper.

\bibliographystyle{unsrt}  
\bibliography{main}  

\end{document}